\documentclass[journal]{IEEEtran} % lettersize
\usepackage{amsmath,amsfonts}
\usepackage{algorithmic}
\usepackage{algorithm}
\usepackage{array}
\usepackage[caption=false,font=normalsize,labelfont=sf,textfont=sf]{subfig}
\usepackage{textcomp}
\usepackage{stfloats}
\usepackage{url}
\usepackage{verbatim}
\usepackage{graphicx}
\usepackage{cite}
\usepackage{siunitx}

\usepackage{booktabs}
\usepackage{footnote}
\usepackage{multirow}
\usepackage{makecell}

\usepackage{hyperref}

\usepackage{soul}

\usepackage{etoolbox}
\makeatletter
\patchcmd{\@makecaption}
  {\scshape}
  {}
  {}
  {}
\makeatother

\begin{document}

\title{Disturbance-Aware Flight Control of Robotic Gliding Blimp via  Moving Mass Actuation}

\author{Hao Cheng and Feitian Zhang$^*$
        % <-this % stops a space
\thanks{This research was in part supported by the National Natural Science Foundation of China under Grant 62473006. 
\textit{(Corresponding author: Feitian Zhang.)}}%
\thanks{The authors are with the Robotics and Control Laboratory, School of Advanced Manufacturing and Robotics, and the State Key Laboratory of Turbulence and Complex Systems, Peking University, Beijing, 100871, China (e-mail: h-cheng@stu.pku.edu.cn; feitian@pku.edu.cn).}
}

\markboth{ }%
{CHENG \MakeLowercase{\textit{et al.}}: Disturbance-Aware Flight Control of Robotic Gliding Blimp via  Moving Mass Actuation} 


\maketitle

\begin{abstract}
Robotic blimps, as lighter-than-air (LTA) aerial systems, offer long endurance and inherently safe operation but remain highly susceptible to wind disturbances. 
Building on recent advances in moving mass actuation, this paper addresses the lack of disturbance-aware control frameworks for LTA platforms by explicitly modeling and compensating for wind-induced effects. 
A moving horizon estimator (MHE) infers real-time wind perturbations and provides these estimates to a model predictive controller (MPC), enabling robust trajectory and heading regulation under varying wind conditions. 
The proposed approach leverages a two-degree-of-freedom (2-DoF) moving-mass mechanism to generate both inertial and aerodynamic moments for attitude and heading control, thereby enhancing flight stability in disturbance-prone environments. 
Extensive flight experiments under headwind and crosswind conditions show that the integrated MHE–MPC framework significantly outperforms baseline PID control, demonstrating its effectiveness for disturbance-aware LTA flight. 
\end{abstract}

\begin{IEEEkeywords}
Mechanisms design and control, flying robots, unmanned autonomous systems. 
\end{IEEEkeywords}

\section{Introduction}
% [Background and motivation]
\IEEEPARstart{S}{mall} aerial robots increasingly demand long endurance, gentle interaction, and safe operation near people and infrastructure. 
Lighter-than-air (LTA) platforms, such as robotic blimps, offer a compelling solution by leveraging static buoyancy to reduce propulsion load, extend flight duration, and lower energy consumption \cite{TmechBlimp1,ManipSurvey,ManipSurvey2,ManipSurvey3}. 
Their applications include search and rescue \cite{searchRescue1,Blimp6}, environmental monitoring \cite{EnvMoni1,EnvMoni2}, human–robot interaction \cite{HRI0,HRI1,HRI2,Blimp7}, and public entertainment \cite{Joy1,Joy2}. 
However, the same buoyancy that enables efficient and safe flight also increases sensitivity to wind disturbances, posing challenges for reliable control. 
Robust real-time estimation and compensation strategies are therefore essential for achieving stable attitude and heading regulation in stochastic, spatially varying airflow environments \cite{RGBlimpQ}. 

% [Mechanism and control gaps in prior work]
Prior work on LTA systems has examined localization, altitude regulation, and partial disturbance rejection \cite{TmechBlimp3,TmechBlimp2,robustBlimp1a,robustBlimp1c,robustBlimp2,robustBlimp3a,Swing,RGBlimp,RAL2024}. 
Although advanced disturbance estimation and adaptive control have substantially improved robustness in multi-rotor platforms\cite{DroneRobust1,DroneRobust2,DroneRobust3,DroneRobust4}, their direct application to blimps remains limited. LTA vehicles exhibit strong aerodynamic coupling, low flight speeds, asymmetric thrust configurations, and buoyancy-dominated dynamics, all of which differ fundamentally from rotary-wing systems. 
These distinctions hinder the transfer of disturbance‑aware control frameworks to robotic blimps and restrict their deployment in realistic windy environments. 
\begin{figure}[t]
      \centering
      \includegraphics[width=0.47\textwidth]{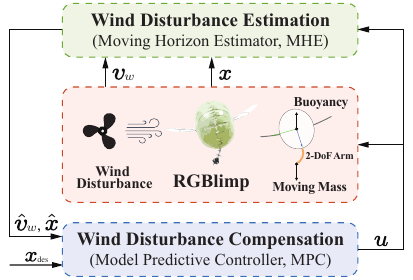}
      \vspace{1em}
      \caption{Schematic of wind disturbance estimation and compensation for RGBlimp with moving mass actuation. A moving horizon estimator provides real-time wind estimates to the model predictive controller, enabling disturbance-aware heading regulation using a 2-DoF moving mass mechanism. } 
      \label{fig.1}
      \vspace{1em}
\end{figure}

\IEEEpubidadjcol

Our prior work \cite{RGBlimpQ} introduced a continuum-based 2-DoF moving mass actuation system for a robotic gliding blimp (RGBlimp), enabling pitch and roll control via gravitational moments and heading regulation via aerodynamic yaw moments. 
This mechanism establishes a foundation for precise and lightweight attitude control. 

Building on this foundation, this paper incorporates environmental wind disturbances directly into the modeling, estimation, and control architecture. 
We develop a moving horizon estimator (MHE) to reconstruct real-time airflow disturbances and integrate these estimates into a model predictive controller (MPC) for disturbance compensation. 
The resulting disturbance-aware control framework significantly improves heading robustness under headwind and crosswind conditions, narrowing the gap between buoyancy-driven platforms and modern disturbance-aware flight control.  
Comprehensive physical flight trials further validate the approach, demonstrating substantial performance gains over open-loop and PID-based controllers. 

The main contributions of this paper are threefold: 1) an integrated MHE-based wind disturbance observer leveraging the nonlinear dynamic model of RGBlimp, 2) a disturbance compensation framework employing MPC-based moving mass actuation, and 3) extensive experimental validation demonstrating enhanced flight robustness against environmental wind disturbances. 

The remainder of this paper is organized as follows. 
Section~\ref{sec.rgblimp} introduces the RGBlimp platform and its actuation mechanisms. 
Section~\ref{sec.model} develops the dynamic model with explicit wind–blimp interaction. 
Section~\ref{sec.control} presents the MHE-MPC disturbance-aware flight control framework.
Section~\ref{sec.ex} reports the experimental results and associated analysis. 
Finally, Section~\ref{sec.conclusion} provides the concluding remarks.

\section{RGBlimp Mechanism and Actuation Principles}
\label{sec.rgblimp}
The RGBlimp studied in this work features an aerodynamic-buoyant hybrid lift design combined with a continuum-based 2-DoF internal moving mass actuator. 
Specifically, fixed wings provide aerodynamic lift, while a helium-filled envelope generates static buoyancy. 
A cable-driven continuum mechanism actuates an internal moving mass at its tip, enabling 2-DoF control to generate gravitational moments for pitch and roll regulation and aerodynamic moments for yaw adjustment. 
This section introduces the RGBlimp system, detailing the continuum-based moving mass mechanism and its role in compensating for wind disturbances to maintain robust flight attitude in such a disturbance-sensitive LTA platform. 

\subsection{Continuum-Based 2-DoF Moving-Mass Actuation}
\begin{figure}[htp]
      \centering
      \includegraphics[width=0.49\textwidth]{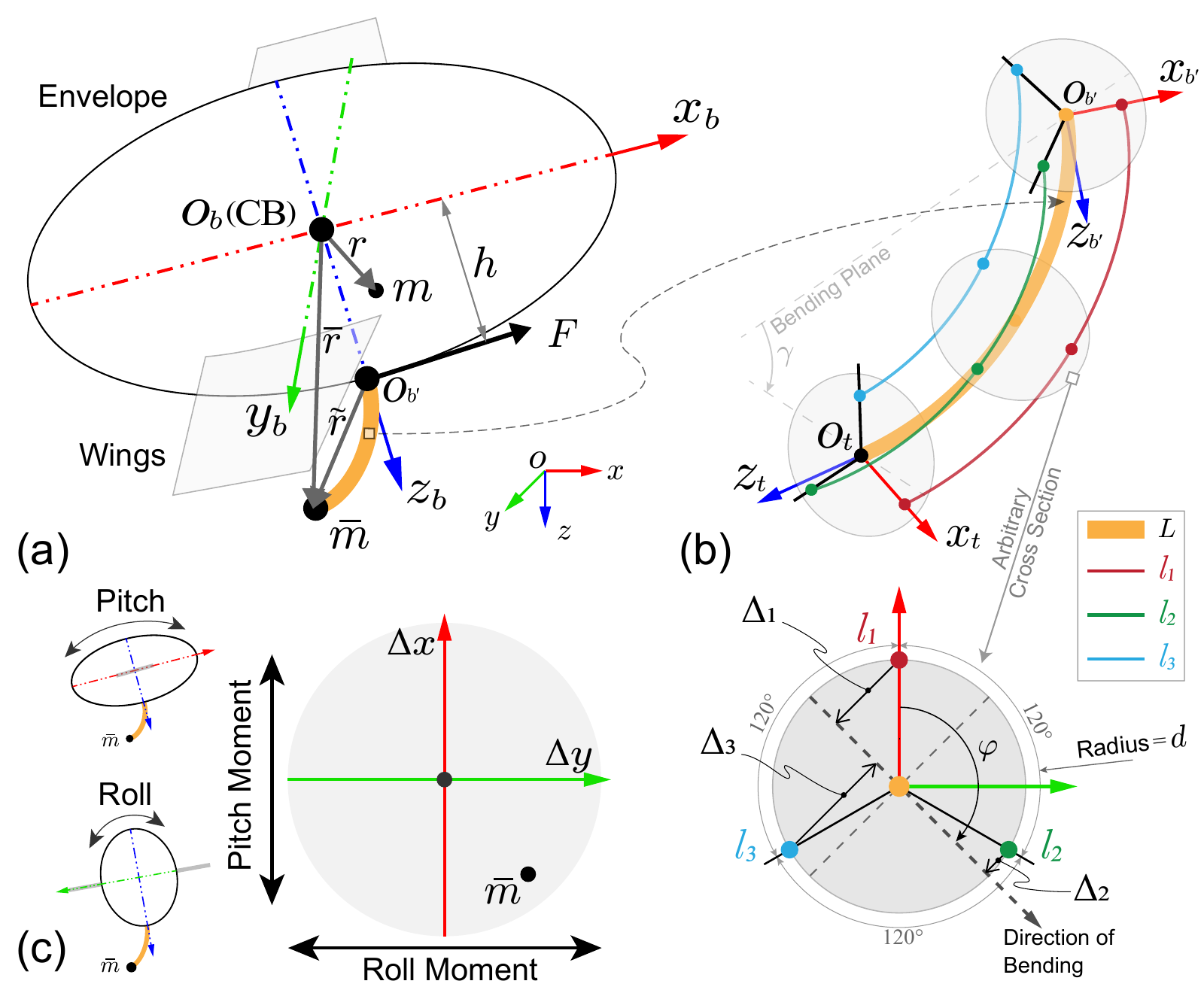}
      \vspace{-2mm}
      \caption{Diagrams of the RGBlimp with  continuum-based internal moving mass. 
               (a) Overall RGBlimp with helium-filled envelope, wings, and moving mass mechanism. 
               (b) Cable-driven continuum mechanism actuating the tip mass $\bar{m}$. 
               (c) Relationship between 2-DoF moving mass and pitch/roll moments. 
               } 
      \label{fig.2}
      \vspace{-0.mm}
\end{figure}

Figure~\ref{fig.2} illustrates the RGBlimp system, highlighting the internal moving mass actuated through a cable-driven continuum arm and the corresponding inertial moments induced by its weight. 
A body-fixed North-East-Down (NED) reference frame, $O_{b}\text{-}x_{b}y_{b}z_{b}$ is established at the center of buoyancy (CB) $O_{b}$. 
The continuum arm base $O_{b^\prime}$ is located directly below the CB at a vertical offset $h$. 
The robot attitude is represented by $\boldsymbol{e}\!=\![\phi,\theta,\psi]^\mathrm{T}$ in the roll-pitch-yaw Euler angle sequence. 

The RGBlimp utilizes a single propeller for translational motion and an internal moving-mass mechanism for attitude control, as illustrated in Fig.~\!\ref{fig.2}(a). 
The propeller, mounted at the bottom, generates thrust along the body-fixed $x_b$ axis. 
Attitude regulation is achieved by an internal moving mass $\bar{m}$, which adjusts the effective center of mass and is actuated by a cable-driven continuum mechanism with 2-DoF motion (Fig.~\!\ref{fig.2}(b)). 
The continuum arm is driven by three cables $l_1$, $l_2$, and $l_3$, and constrained by an incompressible elastic backbone of length $L$. 
By reconfiguring the continuum, the gravitational force of the moving mass generates controllable pitch and roll moments, enabling attitude regulation of the robot (Fig.~\!\ref{fig.2}(c)). 

The continuum arm is modeled as a constant-curvature, untwisted curve with a fixed backbone length---a common assumption in continuum mechanics \cite{ContinuumPCC}. 
Under this assumption, the backbone exhibits constant spatial curvature that evolves over time. 
The transformation from the base frame $O_{b'}\text{-}x_{b'}y_{b'}z_{b'}$ to the tip frame $O_{t}\text{-}x_{t}y_{t}z_{t}$ is determined by the bending direction $\phi\in[-\pi,\pi)$ and curvature angle $\gamma\in[0,\pi/2]$, denoted by $\boldsymbol{T}_{\!b'}^t (\varphi,\gamma)\in\mathrm{SE}(3)$, which gives the translation vector of the continuum arm. 
Considering three uniformly distributed driven cables, the position of the moving mass $\bar{m}$ relative to the CB is denoted by $\bar{\boldsymbol{r}}$ and represented using the $q$-parametrization \cite{RGBlimpQ,QParam}, i.e.,
\begin{equation}
    \bar{\boldsymbol{r}}(\boldsymbol{q}) = [0,0,h]^\mathrm{T} + \frac{Ld}{\delta^2}
    \begin{bmatrix}
      \ \delta_x(1-\cos\frac{\delta}{d})\ \\[+1mm]
      \ \delta_y(1-\cos\frac{\delta}{d})\ \\[+1mm]
        \delta\sin\frac{\delta}{d}
    \end{bmatrix}, 
	\label{eq.1}
\end{equation}
where \(\delta = (\delta_x^2 + \delta_y^2)^{1/2}\), 
scale $d$ is the distance between each cable and the backbone, 
$h$ is the offset of the thrust from the $x_b$ axis, and $\boldsymbol{q}\!=\![\delta_x, \delta_y]^\mathrm{T}$ is defined via cable lengths as
\begin{equation}
    \delta_x  = \tfrac{1}{3}(l_2+l_3-2l_1), \quad
    \delta_y  = \tfrac{\sqrt{3}}{3}(l_3-l_2).\label{eq.2}
\end{equation}

The parameters $\delta_x$ and $\delta_y$ independently modulate the pitch and roll moments, enabling attitude adjustment largely decoupled from aerodynamic effects (Fig.~\!\ref{fig.2}(c)). 
This 2-DoF moving-mass mechanism significantly enhances control authority in the air, which is essential for effective wind-disturbance compensation and robust flight performance. 

\subsection{Heading Regulation via Aerodynamic and Actuation Effects}
\label{sec.heading}
The RGBlimp attitude is represented by the Euler angles---pitch $\theta$, roll $\phi$, and yaw (heading) $\psi$. 
Pitch and roll are controlled via the 2-DoF moving mass inertial moments. 
Heading control is critical for robust flight and is achieved through both aerodynamic and actuation mechanisms. 

\begin{figure}[htp]
      \centering
      \vspace{-0mm}
      \subfloat[]{
      \includegraphics[width=0.23\textwidth]{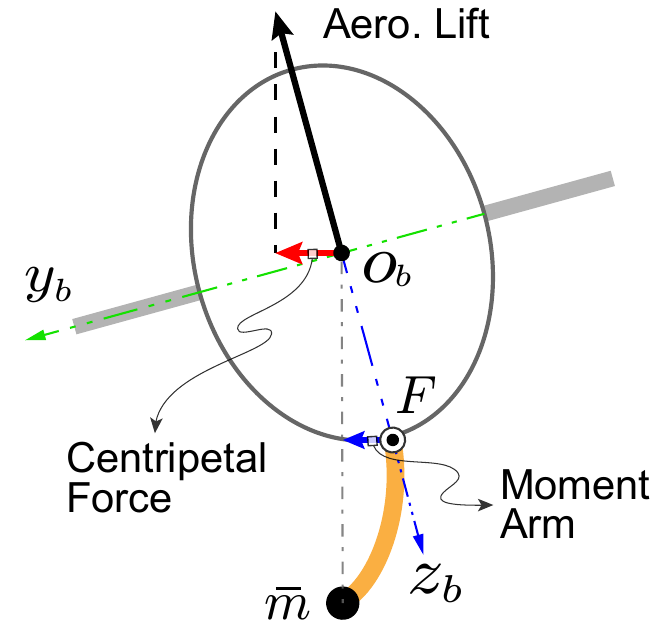}
      } 
      \hspace{-4mm}
      \subfloat[]{
      \includegraphics[width=0.23\textwidth]{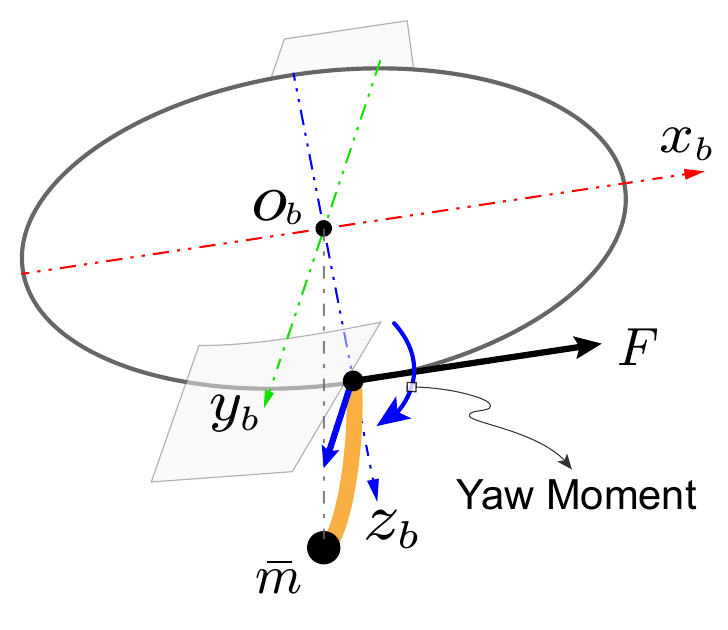}
      }
      \caption{Illustration of the heading adjustment mechanisms. 
               (a) Roll-induced aerodynamic lift produces centripetal force for heading control. 
               (b) Deflection of the continuum arm generates yaw moment via propeller thrust, enabling heading adjustment during low-speed flight. 
      }
      \label{fig.3}
\end{figure}

%\begin{comment}    
%Firstly, consider the aerodynamic yaw moment for heading control. 
%The correlation, generally, between the roll angle and the heading angle in aerial vehicles is essential for accurate maneuvering and disturbance compensating control. 
%When the roll angle is modified, the resulting differential angle of attack (AoA) on the left and right wings induces a lift asymmetry. 
%This asymmetry generates a net lateral force that acts as a centripetal force, thereby producing a yaw moment. 
%Specifically, the wing experiencing an increased angle of attack generates greater lift, while the opposite wing, with a decreased angle of attack, produces less lift. 
%This differential lift facilitates the turning of the RGBlimp by providing the necessary yawing moments, illustrating the critical interplay between roll-induced lift variation and heading control for enhanced maneuverability and stability. 
%\end{comment}
Aerodynamic yaw moments arise from roll-induced lift asymmetry.
Differential angles of attack between the left and right wings generate lateral forces acting as centripetal forces, producing a yaw moment that turns the RGBlimp (Fig.~\ref{fig.3}(a)).

%In addition, consider the yaw moment for heading control that does not rely on aerodynamics, which is crucial for such a low-speed flying robot. 
%When the flight speed is sufficient, such as $\SI{1}{m/s}$, controlling the $\delta_y$ to adjust the roll angle $\phi$ can generate effective aerodynamic yaw moments for heading adjustments. 
At low speeds, aerodynamic yaw moment $\boldsymbol{M}_\mathrm{yaw}$ diminishes as $\left \| \boldsymbol{M}_\mathrm{yaw} \right \|\!\propto\!V^2$, where $V$ is the magnitude of the relative flow speed. 
In this regime, propeller thrust $F$ combined with deflection $\delta_y$ generates yaw moments independently of traveling speed, enabling effective heading control even during slow flight or hovering (Fig.~\ref{fig.3}(b)). 

\section{Dynamics With Disturbance Representation}
\label{sec.model}
Denote the CB position in the inertial frame $O\text{-}xyz$ by $\boldsymbol{p}\in\mathbb{R}^3$. 
The CB translational velocity expressed in the body-fixed frame is $\boldsymbol{v}_b\in\mathbb{R}^3$, and the angular velocity is $\boldsymbol{\omega}_b\in\mathbb{R}^3$. 
The system states are compactly written as $\boldsymbol{x}=[\boldsymbol{p},\boldsymbol{e},\boldsymbol{v}_b,\boldsymbol{\omega}_b]^\mathrm{T}$, and the control inputs are $\boldsymbol{u}=[F,\delta_x,\delta_y]^\mathrm{T}$. 
The RGBlimp dynamics are expressed as
\begin{align}
    \dot{\boldsymbol{p}}=\boldsymbol{R}\boldsymbol{v}_b, \quad\ \ \ \dot{\boldsymbol{R}}=\boldsymbol{R}\boldsymbol{\omega}_b^\times, \nonumber\\[0.5em]
    \boldsymbol{M}(\delta_x,\delta_y)\!
    \begin{bmatrix}
      \,\dot{\boldsymbol{v}}_b\,\\[+0.35mm]
      \,\dot{\boldsymbol{\omega}}_b\,
    \end{bmatrix}
    =
    \begin{bmatrix}
      \,\tilde{\boldsymbol{f}}\,\,\,\\[+1mm]
      \,\tilde{\boldsymbol{t}}\,\,
    \end{bmatrix},\label{eq.3}
\end{align}
\vspace{-2mm}
where
\begin{align}
    % ~f:
    \hspace{1mm}\tilde{\boldsymbol{f}}\hspace{0.5mm} &= (m\!+\!\bar{m})\boldsymbol{v}_b\!\times\!\boldsymbol{\omega}_b+
    (\boldsymbol{\omega}_b\!\times\!\boldsymbol{l}_g)\!\times\!\boldsymbol{\omega}_b\hspace{0.8mm}+\nonumber\\
    &\hspace{4.5mm}(mg\!+\!\bar{m}g\!-\!B)\boldsymbol{R}^\mathrm{T}\boldsymbol{e}_z + \boldsymbol{F}_\mathrm{aero} + F\boldsymbol{e}_x, \label{eq.4} \text{ and}\\[+1.5mm]
    % ~t:
    \hspace{1mm}\tilde{\boldsymbol{t}}\hspace{0.7mm} &= 
    \boldsymbol{l}_g\!\times\!(\boldsymbol{v}_b\!\times\!\boldsymbol{\omega}_b) + 
    \left(\boldsymbol{I}\!-\!\bar{m} (\boldsymbol{\bar{r} }^\times)^2\right)\boldsymbol{\omega}_b\!\times\!\boldsymbol{\omega}_b + \nonumber\\
    &\hspace{5mm}\boldsymbol{l}_g\!\times\!g\boldsymbol{R}^\mathrm{T}\boldsymbol{e}_z + \boldsymbol{T}_\mathrm{aero} +  Fh\boldsymbol{e}_y, \label{eq.5}
\end{align}
%\vspace{-8mm}
\begin{align}
    \boldsymbol{M}(\delta_x,\delta_y)
    &=\!
    %\hspace{3.1mm}
    \begin{bmatrix}
      (m\!+\!\bar{m})\boldsymbol{1}_{3}&  -\boldsymbol{l}_g^\times\\[+0.5mm]
      \boldsymbol{l}_g^\times&  \boldsymbol{I}\!-\!\bar{m} (\boldsymbol{\bar{r} }^\times)^2
    \end{bmatrix}.\quad\quad \label{eq.6}
\end{align}
Here, $\boldsymbol{R}\in\mathrm{SO}(3)$ is the rotation matrix from the body frame to the inertial frame, 
$\boldsymbol{\omega}_b^\times$ stands for the skew-symmetric matrix corresponding to $\boldsymbol{\omega}_b$. 
Scalars $m$, $\bar{m}$, and $B$ represent the stationary mass, moving mass, and buoyancy, respectively. 
$\boldsymbol{l}_g\!=\!m\boldsymbol{r}\!+\!\bar{m}\bar{\boldsymbol{r}}$ is the total center-of-mass vector. 
$\boldsymbol{I}$ represents the moment of inertia of the stationary body. 
$\boldsymbol{1}_3$ is the $3\times3$ identity matrix, and $\boldsymbol{e}_x\!=\![1,0,0]^\mathrm{T}$, $\boldsymbol{e}_y\!=\![0,1,0]^\mathrm{T}$, $\boldsymbol{e}_z\!=\![0,0,1]^\mathrm{T}$ are unit vectors along the body axes. 
%The acceleration of the internal moving mass is neglected due to its minor effect in prototype experiments. 

Considering the wind disturbances, let $\boldsymbol{v}_w\!\in\!\mathbb{R}^3$ denote the environmental wind velocity in the inertial frame. 
The relative velocity is then calculated as $ \boldsymbol{v}_a = \boldsymbol{R}\boldsymbol{v}_b-\boldsymbol{v}_w$.

The aerodynamic forces $\boldsymbol{F}_\mathrm{aero}$ and moments $\boldsymbol{T}_\mathrm{aero}$ are expressed as
\begin{align}
    \boldsymbol{F}_\mathrm{aero}&=\boldsymbol{R}_v^b\,[-D,\,S,\,-L]^\mathrm{T}, \text{ and}\\
    \boldsymbol{T}_\mathrm{aero}&=\boldsymbol{R}_v^b\,[T_x,\,T_y,\,T_z]^\mathrm{T},\label{eq.7}
\end{align}
with $\boldsymbol{R}_v^b=\boldsymbol{R}_y(-\alpha)\boldsymbol{R}_z(\beta)$ where $\alpha = \arctan(v_{az}/v_{ax})$ and $\beta = \arcsin(v_{ay}/V_a)$.

\noindent Forces $[D, S, L]$ and moments $[T_x, T_y, T_z]$ are modeled as 
\begin{align*}
    [D,\,S,\,L]^\mathrm{T}&=\,\tfrac{1}{2}\rho V^2\!A[C_D, C_S, C_L]^\mathrm{T}, \text{ and}\\
    [T_{\!x},T_{\!y},T_{\!z}]^\mathrm{T}\!\!&=\!\tfrac{1}{2}\rho V^2\!\!A[C_{\!T_{\!x}}, C_{\!T_{\!y}}, C_{\!T_{\!z}}]^\mathrm{T}\!\!+\![K_{\!x},K_{\!y},K_{\!z}]^\mathrm{T}\,\boldsymbol{\omega}_b,
\end{align*}
where \!$A$ is the aerodynamic reference area, and coefficients denoted by ``$C$'' and ``$K$'', which are dependent on $(\alpha,\beta)$, capture aerodynamic and damping effects \cite{RGBlimp,RGBlimpQ}.

\section{Wind Estimation and Motion Control}
\label{sec.control}
%This section presents a moving-horizon estimator and model predictive control to maintain stable flight in complex airflow conditions.

\subsection{Moving Horizon Estimation of Wind Velocity}
\label{sec.mhe}
The moving horizon estimator (MHE) provides real-time, joint state-parameter estimation for nonlinear dynamical systems by solving a constrained optimization problem over a finite receding horizon. 
For the convenience of derivation, we discretize the RGBlimp dynamics (Eq.\!~\eqref{eq.3}) as $\boldsymbol{x}_{i+1}=\boldsymbol{f}(\boldsymbol{x}_i,\boldsymbol{u}_i,\boldsymbol{v}_w,\Delta t)$, where $\boldsymbol{v}_w$ denotes the unknown wind disturbance treated as a time-invariant or slowly varying parameter over the horizon, and $\Delta t$ is the sampling time. % used in the iterative optimization process. 
At each time step $t\!\geq\!N$, the MHE estimator optimizes over $N\!+\!1$ states \mbox{$\mathbf{x}\!=\!\{\boldsymbol{x}_{t-N},\ldots,\boldsymbol{x}_{t}\}$} and a time-invariant wind disturbance $\boldsymbol{v}_w \in \mathbb{R}^3$ on the basis of $N\!+\!1$ measurements \mbox{$\mathbf{y}\!=\!\{\boldsymbol{y}_{t-N},\ldots,\boldsymbol{y}_{t}\}$} and $N$ control inputs \mbox{$\mathbf{u}\!=\!\{\boldsymbol{u}_{t-N},\ldots,\boldsymbol{u}_{t-1}\}$}. 
To infer $\boldsymbol{v}_w$ jointly with the states, we formulate the MHE as the following constrained nonlinear program, i.e., 
\begin{align}
    \displaystyle
    &\!\min_{\mathbf{x}_{t-N\hspace{-0.1mm}:\hspace{-0.3mm}t}\!, \boldsymbol{v}_w}\,\!\left \| \boldsymbol{x}_{t-N} - \hat{\boldsymbol{x}}_{t-N}  \right \|_{\boldsymbol{P}\!_x}^2 + \!\left \| \boldsymbol{v}_w - \hat{\boldsymbol{v}}_w  \right \|_{\boldsymbol{P}\!_w}^2 \quad\nonumber\\[-0.5mm]
    &\quad\quad\quad\!+\!{\textstyle\sum_{i=t-N}^{t}} \left \| \boldsymbol{y}_i - \boldsymbol{h}(\boldsymbol{x}_i)  \right \|_{\boldsymbol{P}_{\!u}}^2 \nonumber\\[1mm]
    &\ \ \mathrm{s.t.}\ \, \boldsymbol{x}_{i+1}=\boldsymbol{f}(\boldsymbol{x}_i,\boldsymbol{u}_i,\boldsymbol{v}_w,\Delta t). \label{eq.22} %\quad i = 0,1,...,N\!-\!1. 
\end{align}
Here, $\hat{\boldsymbol{x}}_{t-N}$ and $\hat{\boldsymbol{v}}_w$ are the MHE estimates, and $\boldsymbol{h}(\cdot)$ denotes the measurement map depending on states, inputs, and parameters. 
For any vector $\boldsymbol{z}$, $\left \|\boldsymbol{z}\right \|_{\boldsymbol{P}}^2\!=\!\boldsymbol{z}^\mathrm{T}\!\boldsymbol{P}\boldsymbol{z}$ denotes the weighted squared norm with $\boldsymbol{P}$ being positive definite.  
%At each step, the optimization yields the estimated trajectory $\hat{\boldsymbol{x}}_{0\hspace{-0.1mm}:\hspace{-0.3mm}N}$ and the estimated wind velocity ${\boldsymbol{v}}_w$. 
%${\boldsymbol{v}}_w$ is assumed to be constant over the estimation horizon. 

The cost function consists of three components: 
1) the arrival cost, summarizing information prior to the current horizon; 
2) the disturbance cost, penalizing the deviation of the estimated wind from its prior; and 
3) the measurement running cost, enforcing consistency between measured and predicted outputs. 

\subsection{MPC Design for Moving Mass Control}
\label{sec.mpc}
Given the estimated wind disturbance and the RGBlimp dynamics, the model predictive control (MPC) computes the optimal actuation sequence to maintain a robust heading in complex wind fields. 
We define the finite-horizon optimal control problem with a prediction window of $M$ steps as 
\begin{align}
    \displaystyle
    &\min_{\substack{\mathbf{x}_{t\hspace{-0.1mm}:\hspace{-0.0mm}t+\!M}\\[0.2mm] \mathbf{u}_{t\hspace{-0.1mm}:\hspace{-0.0mm}t+\!M\!-\!1}}}\,\sum_{j=t}^{t+\!M} \left \| \boldsymbol{x}_j - \boldsymbol{x}_{\mathrm{des},j}  \right \|_{\boldsymbol{Q}_{\!x}}^2 
    + \sum_{j=t+1}^{t+\!M\!-\!1} \left \| \boldsymbol{u}_j - \boldsymbol{u}_{j\!-\!1}  \right \|_{\boldsymbol{Q}_{\!u}}^2\nonumber\\[1mm]
    &\ \ \mathrm{s.t.}\ \ \ \boldsymbol{x}_t=\hat{\boldsymbol{x}}_t, \text{ and}\nonumber\\
    &\hspace{10.8mm}\boldsymbol{x}_{j+1}=\boldsymbol{f}(\boldsymbol{x}_j,\boldsymbol{u}_j,\hat{\boldsymbol{v}}_w,\Delta t). \label{eq.23}%,\quad j = 0,1,...,M\!-\!1
\end{align}
Here, $\boldsymbol{Q}_{\!x}$ and $ \boldsymbol{Q}_{\!u}$ are positive-definite weighting matrices. 
The control sequence $\mathbf{u}_{t\hspace{-0.1mm}:\hspace{-0.0mm}t+\!M\!-\!1}\!=\!\{\boldsymbol{u}_{t},\ldots,\boldsymbol{u}_{t+\!M-1}\}$ determines the actuation input over the prediction window. 
The predicted states $\{\boldsymbol{x}_{j}\}_{j=t}^{t+\!M}$ are generated from the dynamic model $\boldsymbol{f}(\cdot)$ using the current state estimate $\hat{\boldsymbol{x}}_t$ and the disturbance estimate  $\hat{\boldsymbol{v}}_w$, which is assumed to be constant over the prediction horizon. 
The discrepancy term $\boldsymbol{x}_j\!-\!\boldsymbol{x}_{\mathrm{des},j}$ denotes the deviation from the desired state trajectory, while the control-difference penalty ensures smooth actuation without aggressive changes. 
In this study, the primary objective is robust heading regulation, implemented by prescribing a constant desired yaw, i.e., $\psi_{\mathrm{des},j}\equiv\psi_{\mathrm{const}}, \forall j$. 
In addition, we impose a desired body angular velocity $\boldsymbol{\omega}_{b,\mathrm{des}}\equiv0$, which suppresses excessive roll and pitch motions, enhancing stability under wind disturbances.

\subsection{Weighting Matrix Design}
The performance of both the MHE and MPC frameworks critically depends on the selection of weighting matrices, which govern the trade-offs among estimation accuracy, control effort, and state regulation. 
In this work, we employ diagonal weighting matrices and assign their entries according to the physical units, noise characteristics, and relative importance of each state and input.
The weighting matrices used for the RGBlimp platform are
\begin{align}
    \boldsymbol{P}\!_x &= \mathrm{diag}(\underbrace{10,10,10}_{\text{Position }\boldsymbol{p}}; \underbrace{10,10,10}_{\text{Attitude }\boldsymbol{e}}; \underbrace{10,10,10}_{\text{Velocity }\boldsymbol{v}_b}; \underbrace{10,10,10}_{\text{Angular Vel. }\boldsymbol{\omega}_b}),\nonumber\\
    \boldsymbol{P}\hspace{-0.95mm}_w &= \mathrm{diag}(50,50,50),\nonumber\\
    \boldsymbol{P}\!_u &= \mathrm{diag}(5,5,5);\nonumber\\
    \boldsymbol{Q}\hspace{-0.2mm}_x &=\mathrm{diag}(0,10,10;0,0,100;0,0,0;5,5,5),\nonumber\\
    \boldsymbol{Q}\hspace{-0.2mm}_u &= \mathrm{diag}(200,300,300). \nonumber
\end{align}
Here, the notation $\mathrm{diag}(\cdot)$ denotes a diagonal matrix. 

In the MHE formulation, the matrix $\boldsymbol{P}\!_x$ applies uniform, moderate weights to all state components to ensure reliable initialization without biasing specific states. 
The disturbance weight $\boldsymbol{P}\!_w$ is set significantly higher than the measurement-noise weight $\boldsymbol{P}\!_u$, prioritizing accuracy in the estimated wind over exact measurement fit, which is essential under modeling uncertainties and sensor noises.

In the MPC design, the tuning reflects the control objectives of robust heading regulation and lateral stability. 
The state-weight matrix $\boldsymbol{Q}\hspace{-0.2mm}_x$ imposes a dominant penalty on yaw error $\psi$ to enforce precise heading control. 
Moderate weights on lateral position $p_y$ and altitude $p_z$ preserve path accuracy, while weights on angular velocity $\boldsymbol{\omega}_b$ limit aggressive rotational motion and promote smoother trajectories. 
Forward position $p_x$ and translational velocity $\boldsymbol{v}_b$ are intentionally unweighted, allowing the controller flexibility in adjusting forward speed under wind disturbances.

The input-weight matrix $\boldsymbol{Q}\hspace{-0.2mm}_u$ strongly penalizes control effort, with higher weights on moving-mass commands ($\delta_x$, $\delta_y$) than on thrust $F$. This choice discourages abrupt internal actuation, reduces actuator wear, and mitigates oscillatory behavior. 
All weights were systematically tuned through simulation and validated in physical experiments to achieve a balance between performance and robustness across diverse wind conditions. 
 
\section{Experiments}
\label{sec.ex}
\subsection{Experimental Setup}
\begin{figure}[t]
      \centering
      \vspace{-0mm}
      \subfloat[]{
      \hspace{-3mm}
      \includegraphics[scale=0.315]{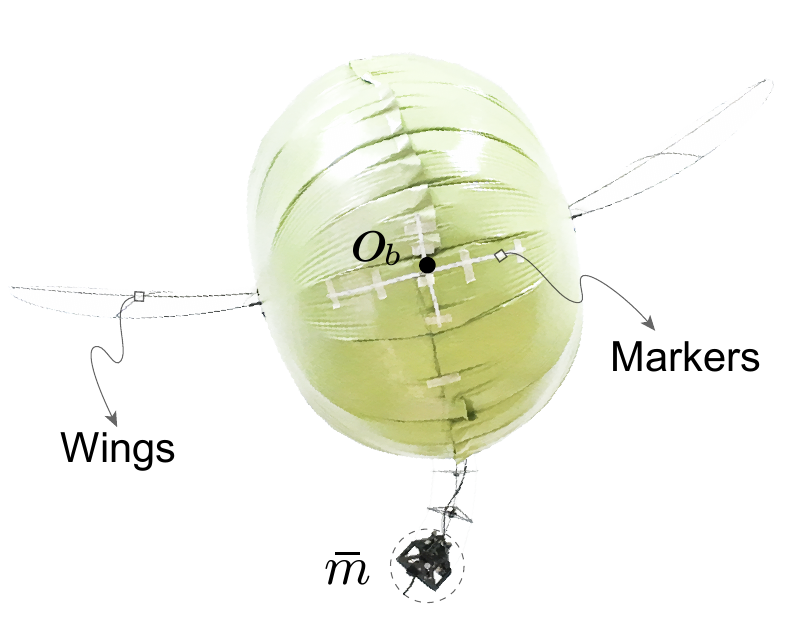}
      } 
      \hspace{-3mm}
      \subfloat[]{
      \includegraphics[scale=0.315]{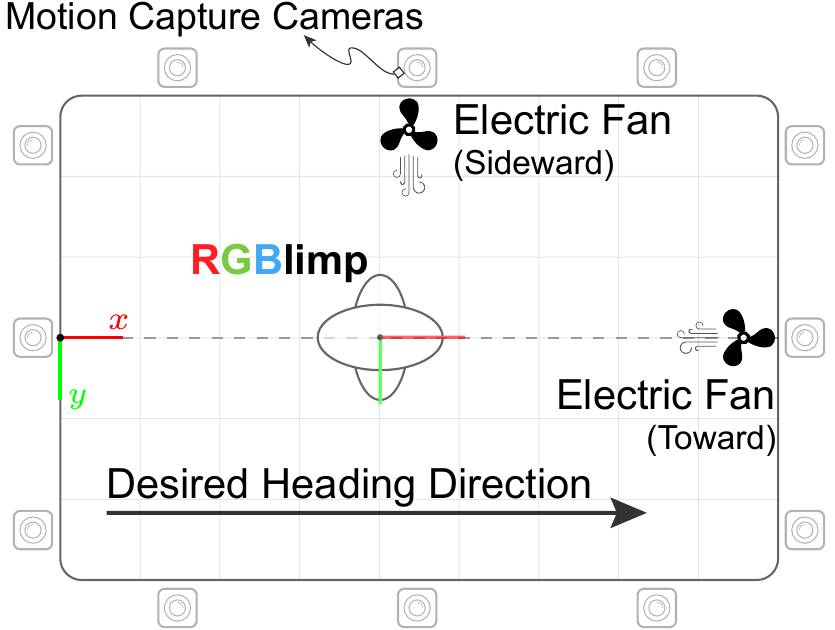}
      }
      \caption{RGBlimp prototype and flight arena. 
               The workspace utilized $x$-axis alignment for maximum flight length tracking by the OptiTrack system. 
              }
      \label{fig.4}
\end{figure}

\textit{System and Prototype:} 
As described in Sec.~\!\ref{sec.rgblimp}, the \mbox{RGBlimp} is a lighter-than-air aerial system characterized by high sensitivity to environmental disturbances and nonlinear aerodynamic effects, making wind compensation particularly challenging. 
To evaluate the proposed MHE-MPC scheme, we employ an RGBlimp prototype \cite{RGBlimpQ} equipped with the continuum-based moving mass actuation mechanism (Fig.~\!\ref{fig.4}(b)). 
The platform has a net weight of \SI{6.7}{gf}, with the helium buoyancy $B\!=\!\SI{194.2}{gf}$ nearly counterbalancing the moving mass $\bar{m}\!=\!\SI{92.2}{gf}$ and the stationary mass $m\!=\!\SI{108.7}{gf}$.
%The parameters of the prototype are detailed in the appendix, including basic parameters and the approximate functions of aerodynamic coefficients. 

\textit{Disturbance and Objectives:}
We assess performance under wind disturbances generated by a \textit{Dyson AM07} electric fan, producing airflow of approximately $\SI{1.5}{m/s}$, directed either head-on or laterally relative to the flight path. 
%The robot is commanded to maintain a consistent heading at $\SI{1.5}{m}$ altitude. 
The primary control objective is robust heading regulation under wind disturbances, with angular-rate suppression for smooth maneuvering. 
The flight altitude is set at $\SI{1.5}{m}$, and the forward body-frame velocity $v_{b,x}$ is constrained below $\SI{1.5}{m/s}$. 
Accordingly, the desired states in Eq.~\!\eqref{eq.23} are 
\begin{align}
    p_{\mathrm{des},y} = 0,\ p_{\mathrm{des},z} = 1.5\,[\SI{}{m}],\ \psi_\mathrm{des} = 0,\ \boldsymbol{\omega}_{b,\mathrm{des}}=\boldsymbol{0}.  \nonumber
\end{align}

\textit{Estimation and Control:} 
During flight, wind disturbances are estimated using the MHE described in Sec.~\!\ref{sec.mhe} and compensated by the MPC in Sec.~\!\ref{sec.mpc}. 
Both optimization problems are solved on a host computer equipped with an Intel Core i7-12700 CPU. 
At each control cycle, the computed optimal inputs are transmitted over Wi-Fi to an onboard ESP32 microcontroller running MicroROS. 

The MHE \eqref{eq.22} and MPC \eqref{eq.23} employ $N\!=\!M\!=\!20$ shooting steps, corresponding to a $\SI{0.5}{s}$ horizon. 
The resulting nonlinear programs are generated using \texttt{CasADi} and solved via sequential quadratic programming using the \texttt{IPOPT} solver through the \texttt{do-mpc} Python toolbox\cite{dompc}. 

State and input constraints are enforced as 
\[
\begin{array}{r@{\;}c@{\;}c@{\;}c@{\;}l@{\;}l@{\;}r@{\;}c@{\;}c@{\;}c@{\;}l@{\;}l}
    -90 & \leq & \boldsymbol{e} & \leq & 90, &\quad [\SI{}{deg}], \hspace{3em}
     0 & \leq & \boldsymbol{v}_b & \leq & 1.5, &\quad [\SI{}{m/s}],\\[0.2mm]
     1 & \leq & F & \leq & 15, &\quad [\SI{}{gf}], \hspace{3em}
     -45 & \leq & \boldsymbol{q} & \leq & 45, &\quad [\SI{}{mm}],
\end{array}
\]
where $\boldsymbol{q}=[\delta_x,\delta_y]^\mathrm{T}$ denotes the continuum-mechanism actuation parameters defined in Eq.~\eqref{eq.2}.

\begin{figure*}[thpb]
      \centering
      \includegraphics[width=1\textwidth]{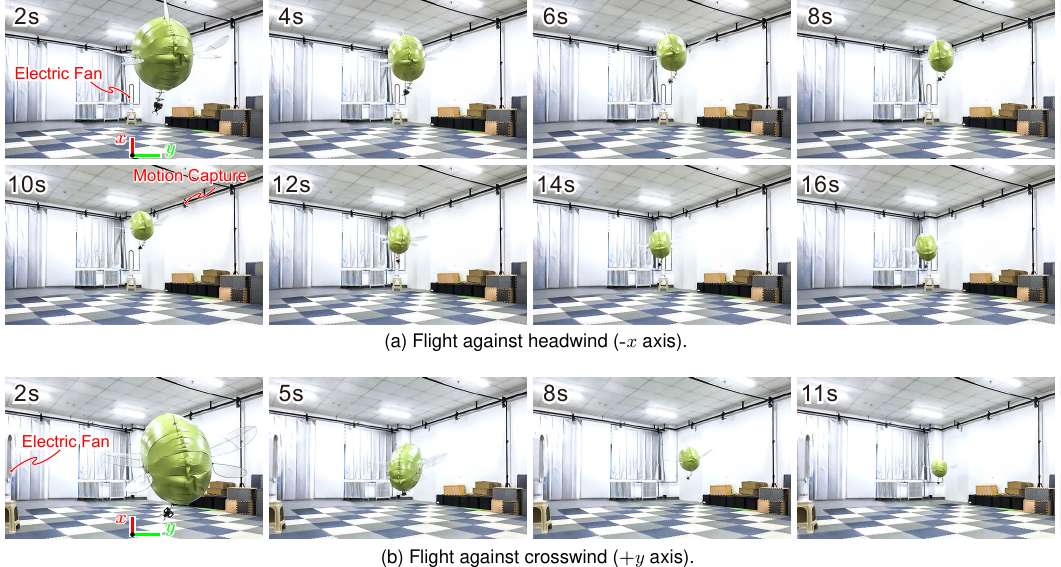} 
      \caption{Flight snapshots demonstrating robust heading control with simultaneous wind estimation and compensation under (a) headwinds and (b) crosswinds. 
               (a) The fan generated a headwind that strengthened from 2 to 8 seconds. 
               The robot advanced into the airflow until the aerodynamic lift balanced its net weight (\SI{6.85}{g}), achieving near-stationary hover between 10 and 16 seconds. 
               (b) The fan produced a crosswind from 5 to 8 seconds. 
               The controller rapidly adjusted the moving-mass configuration to counteract lateral drift, preserving robot's heading and forward progression. } 
      \label{fig.5}
\end{figure*}

\textit{Localization:} 
All experiments were conducted in an \textit{OptiTrack} motion capture environment operating at $\SI{60}{Hz}$ with $\SI{0.8}{mm}$ RMS positional accuracy. 
Twelve onboard infrared LEDs (\SI{850}{nm}) served as active markers, enabling real-time pose estimation. 
Tracking data were streamed via VRPN and ROS2 to support MHE and MPC computation.

\textit{Launch:}
An electromagnetic releaser ensured consistent initial conditions across trials by maintaining a fixed starting pose. 
Upon release, the propeller generated $\SI{10}{gf}$ of thrust, providing sufficient initial speed before activating real-time disturbance estimation and compensation, thereby avoiding excessive angles of attack and mitigating stall risks during the take-off transient. 

%\textit{Metrics:}
%The root mean squared error (RMSE) and the cumulative RMSE (CRMSE) are introduced to evaluate our approach to wind disturbance compensation, where the error metrics between the reference values $\boldsymbol{\xi}$ and the experimental values $\hat{\boldsymbol{\xi}}$ are formulated as
%\begin{align}
%    \text{RMSE}(\boldsymbol{\xi},\hat{\boldsymbol{\xi}})&=\sqrt{\tfrac{1}{n}{\textstyle\sum_{k=0}^{n-1}}(\xi-\hat{\xi}_k)^2},\nonumber\\[1mm]
%    \text{CRMSE}(\boldsymbol{\xi}_T,\hat{\boldsymbol{\xi}}_T)&={\textstyle\sum_{t=0}^{T}}\text{RMSE}(\boldsymbol{\xi}_t,\hat{\boldsymbol{\xi}}_t),\nonumber
%\end{align}
%where $n$ denotes the number of samples, and $T$ represents the number of control iterations from the start of the flight test. 
%The CRMSE characterizes the trend of the error with respect to flight time, making it suitable for evaluating the effectiveness of disturbance compensation control over the entire flight period. 

\subsection{Experimental Results}
Robust flight under wind disturbances is particularly challenging for the RGBlimp due to the stochastic nature of airflow and the aerodynamic sensitivity inherent to lighter-than-air (LTA) vehicles. 
We evaluate the proposed wind-disturbance estimation and compensation framework across diverse scenarios with varying wind intensities and directions. 
In all trials, the RGBlimp is tasked with maintaining a consistent heading at approximately $\SI{1.5}{m}$ altitude under either headwind or crosswind conditions.
Controlled disturbances were generated by placing an electric fan at either in front of the flight path \mbox{$(x\!=\!\SI{6.0}{m},\, y\!=\!\SI{0.0}{m})$} or the side \mbox{$(x\!=\!\SI{3.0}{m},\, y\!=\!\SI{-2.0}{m})$}, as illustrated in Fig.\!~\ref{fig.4}(b). 
Three wind intensities were tested, including no wind ($\SI{0.0}{m/s}$), light wind ($\sim\SI{0.5}{m/s}$), and strong wind ($\sim\SI{1.0}{m/s}$), measured at $\SI{2}{m}$ distance and $\SI{1.5}{m}$ height in front of the fan. 
Because of the fan’s characteristics, the airflow formed a bar-shaped region that attenuated with distance and transitioned from near-laminar to turbulent flow. 

For each condition, ten flight trials were conducted using the proposed disturbance estimation and compensation framework, alongside open-loop trials using constant input $F\!=\!\SI{10}{gf}$ and $\delta_x,\delta_y\!=\!\SI{0}{mm}$ for comparison, resulting in a total of 120~flight trials. % (3*2)*(10*2)=120
Figure~\ref{fig.5} illustrates representative snapshots of the  headwind (Fig.\!~\ref{fig.5}(a)) and crosswind (Fig.\!~\ref{fig.5}(b)) experiments.

The experimental results are presented in two parts to analyze the spatial distribution of wind disturbances and MHE performance, as well as the cumulative RMSE and MPC effectiveness. 
Specifically, the results include:
1) spatial variation of heading angle $\psi$, lateral offset $y$, and estimated disturbance $\boldsymbol{v}_{w}$ along the forward path, and   
2) temporal behavior of heading angle $\psi$, lateral offset $y$, and control inputs $\boldsymbol{u}\!=\![F,\delta_x,\delta_y]^\mathrm{T}$.

\begin{figure*}[p]
      \centering
      \hspace{-0.8em}\includegraphics[width=0.98\textwidth]{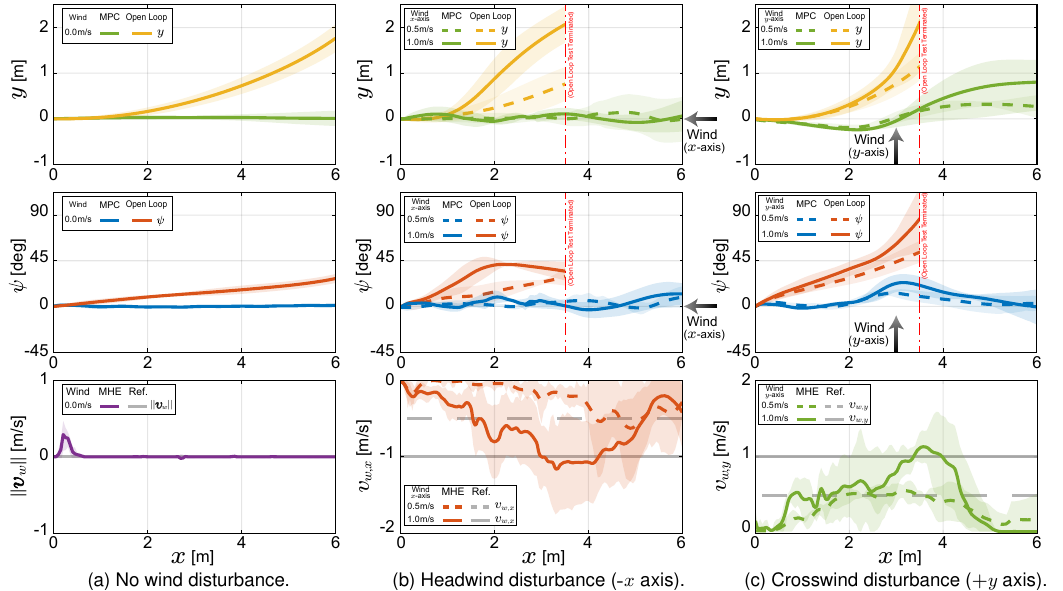} 
      \vspace{-0.6em}
      \caption{The flight experimental results along the forward path. 
               Comparing with MHE-MPC and open loop, plots showing the mean (lines) and the standard deviation (shaded area) of the offset $y$, the heading direction $\psi$, and the estimated wind disturbance $\boldsymbol{v}_{w}$ from five flight trails for three different wind disturbance levels of $\SI{0.0}{m/s}$ (no wind, the first column), $\SI{0.5}{m/s}$ (light wind, dotted lines in the second and third column), and $\SI{1.0}{m/s}$ (strong wind, dotted lines in the second and third columns) and two different directions of $-x$ axis (headwind $v_{w,x}\!<\!0$, the second columns) and $+y$ axis (crosswind $v_{w,y}\!>\!0$, the third column). 
               Note that the flight test is terminated when the the $y$-axis offset distance $y$ reaches limit of $\SI{2}{m}$. 
               For example, open-loop tests under windy conditions are terminated early (dash-dotted vertical lines). } 
      \label{fig.6}

      \vspace{1.2em}
      \includegraphics[width=0.98\textwidth]{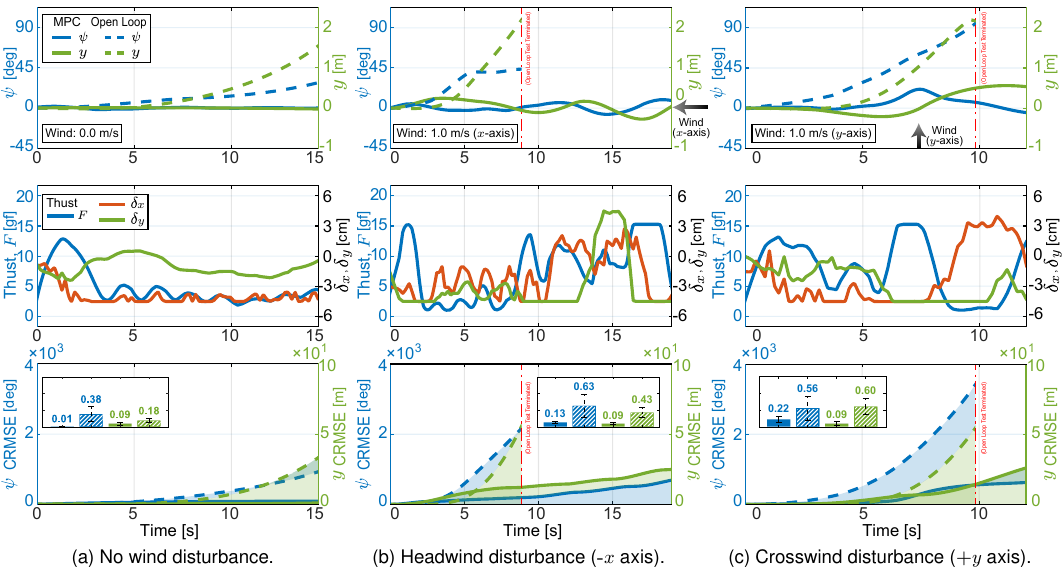} 
      \vspace{-0.6em}
      \caption{The flight experimental results over time. 
               Comparing with MHE-MPC (solid lines) and open loop (dashed lines), plots showing the offset $y$, the heading direction $\psi$, and the control input $\boldsymbol{u}\!=\![F,\delta_x,\delta_y]^\mathrm{T}$ of a certain corresponding flight test with or without wind disturbance in two different directions, that is, $\SI{0.0}{m/s}$ (no wind, the first column) or $\SI{1.0}{m/s}$ (strong wind, the second and third columns), and $-x$ axis (headwind $v_{w,x}\!<\!0$, the second columns) or $+y$ axis (crosswind $v_{w,y}\!>\!0$, the third column). 
               The third row shows the cumulative RMSE over flight time to validate the effectiveness of wind disturbance compensation control, with an inset bar chart presenting the average RMSE over the entire flight. 
               Note that the flight test is terminated when the the \mbox{$y$-axis} offset distance $y$ reaches limit of $\SI{2}{m}$. 
               For example, open-loop tests under windy conditions are terminated early (dash-dotted vertical lines). } 
      \label{fig.7}
\end{figure*}

\textit{1) Spatial Variation and Estimation of Environmental Wind Disturbances}: 
Accurate real-time estimation is critical for effective disturbance-aware control. 
We evaluate the MHE's estimation performance under fan-generated airflow up to $\SI{1}{m/s}$ from either the front (headwind) or side (crosswind). 

Figure~\ref{fig.6} presents the experimental results as a function of forward distance $\boldsymbol{x}$, ranging from $\SI{0}{m}$ to $\SI{6}{m}$. 
The first two rows show the lateral offset $y$ and heading $\psi$, respectively; the third presents the estimated wind disturbance along the flight path or the $x$ axis. 
Flight trials include no wind ($\SI{0.0}{m/s}$), headwind, and crosswind conditions with disturbances of approximately $\SI{1.0}{m/s}$ along the $-x$ and $y$ axes. 
The results show that the MHE consistently reconstructs spatially varying wind disturbances in real time. 
Under headwind conditions, the estimated disturbance $v_{w,x}$ at $x=\SI{4}{m}$ achieves mean values of \SI{0.3}{m/s} and \SI{1.1}{m/s} for nominal wind speeds of \SI{0.5}{m/s} and \SI{1.0}{m/s}, respectively.  
In crosswind scenarios, the estimated $v_{w,y}$ at $x=\SI{3}{m}$ exhibits mean values of \SI{0.5}{m/s} and \SI{0.9}{m/s} under the same nominal wind conditions.  
These results show close agreement with anemometer measurements taken at corresponding locations, thereby validating the effectiveness of the MHE approach. 

The experimental results also reveal that minor machining deviations induced systematic open‑loop flight deflections even in windless conditions (Fig.\!~\ref{fig.6}(a)). 
Under headwind and crosswind conditions, the trial‑to‑trial fluctuations became more pronounced (Fig.\!~\ref{fig.6}(b)(c)), reflecting the inherent randomness of indoor airflow. 
These disturbances exacerbated the deviations observed in open‑loop tests, ultimately driving the robot beyond the experimental boundary ($y\!>\!2$) at approximately \SI{3.5}{m} along the $x$-axis, which necessitated the termination of the trial. 
The comparison with open-loop trajectories demonstrates the effectiveness and repeatability of the proposed MPC-based compensation. 
During takeoff (from $\SI{0}{m}$ to $\SI{0.5}{m}$), the robot experiences large angle of attack (AoA) conditions, causing dynamic model mismatch and transient estimation inaccuracy, which is clearly visible in the fan-off case (Fig.~\!\ref{fig.6}(a)). 

\begin{figure*}[thpb]
      \centering
      \includegraphics[width=1\textwidth]{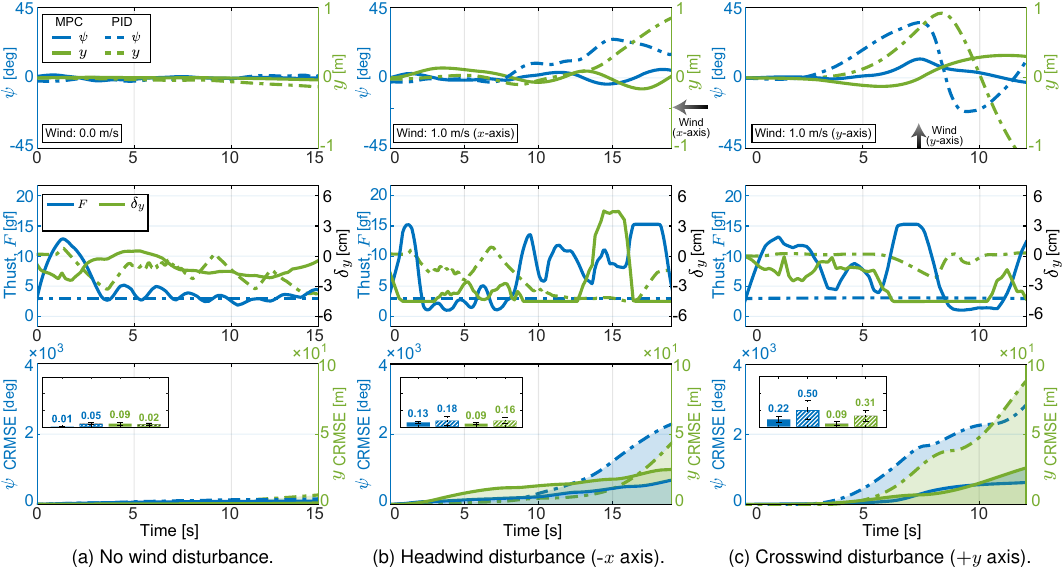} 
      \vspace{-1em}
      \caption{Flight test results comparing a classical model-free PID controller and the proposed MHE-MPC controller. 
               Comparing with MPC (solid lines) and PID (dashed lines), plots showing the offset $y$, the heading direction $\psi$, and the control inputs $F$ and $\delta_y$ (the primary yaw-control input) of a certain corresponding flight test with no wind ($v_{w}\approx\SI{0.0}{m/s}$, the first column), headwind ($v_{w,x}\approx\SI{-1.0}{m/s}$, the second column), and crosswind disturbances ($v_{w,y}\approx\SI{1.0}{m/s}$, the third column). 
               The third row shows the cumulative RMSE over flight time, comparing wind disturbance compensation between our approach and the PID controller, with a bar chart inset displaying the average RMSE across the entire flight. 
               } 
      \label{fig.8}
\end{figure*}

\textit{2) Temporal Variation and Compensation of Environmental Wind Disturbances}: 
Robust forward flight in the presence of headwind or crosswind is difficult due to stochastic airflow and nonlinear, buoyancy-dominated dynamics.
We evaluate disturbance compensation under headwinds and crosswinds of approximately \SI{1.0}{m/s}. 

Figure~\ref{fig.7} illustrates the experimental results of flight trajectories, including heading $\psi$, lateral offset $y$, control inputs $F$, $\delta_x$, $\delta_y$, and cumulative root mean square error (RMSE) over time in $\psi$ and $y$. Scenarios include no wind (first column), headwind, and crosswind disturbances (second and third columns). 
With no wind (fan off), open-loop flights gradually drift due to unmodeled aerodynamic effects. 
In contrast, the MHE-MPC controller compensates for these effects and reduces tracking error by approximately $\SI{90}{\%}$ relative to the open loop. 

Under headwind and crosswind, the RGBlimp undergoes significant drag and lateral forces. 
Open-loop control fails to maintain heading and often reaches the lateral boundary early. 
In contrast, the proposed control actively rejects disturbances, preserving heading and offset stability. 
We observe that the input $\delta_y$ (green lines, Fig.~\ref{fig.7}, second row) plays a key role---inducing roll-generated aerodynamic yaw moments (Sec.~\ref{sec.heading}) to counteract wind-induced yaw.

\subsection{Comparison Against PID Control}
Finally, we compare the proposed MHE-MPC controller with a classical PID baseline. 
Figure~\ref{fig.8} illustrates flight results under no-wind, headwind, and crosswind conditions. 
Across all scenarios, the model-based MHE-MPC consistently outperforms PID, achieving smaller lateral offsets and heading deviations, faster disturbance rejection, and smoother, more coordinated control actions. 
In addition, MHE-MPC yields markedly lower cumulative RMSE for both offset $y$ and heading $\psi$, and a reduced risk of premature termination under strong disturbances. 
The cRMSE obtained after the flight test using the MHE-MPC approach is more than half lower than that achieved by the PID baseline. 
The proposed approach achieves a final $y$-axis offset of \SI{0.05}{m} under headwind conditions and \SI{0.31}{m} under crosswind conditions, significantly outperforming the PID controller, which yields offsets of \SI{0.85}{m} and \SI{-1.13}{m}, respectively. 

We conjecture that these performance gains stem from two key factors.
First, MHE provides recursive state and wind-disturbance estimates, enabling accurate disturbance compensation.  
Second, MPC exploits predictive, constrained optimization to coordinate 2-DoF moving-mass actuation and aerodynamic coupling while avoiding aggressive inputs. 
In contrast, PID lacks disturbance estimation and prediction, leading to locally reactive, suboptimal responses and larger errors under wind.

\section{Conclusion}
\label{sec.conclusion}
This paper presented, to the best of authors' knowledge, the first unified framework that achieved simultaneous wind-disturbance estimation and compensation for a robotic gliding blimp employing 2-DoF moving mass actuation. 
Building on a first-principles dynamic model, the framework employed a moving horizon estimator (MHE) to infer wind disturbances in real time, and a model predictive controller (MPC) that exploited the rapid attitude-regulation capability of the moving mass mechanism to counteract wind-induced deviations. 
Extensive flight experiments under headwind and crosswind conditions were conducted. Experimental results validated the proposed approach, demonstrating accurate wind estimation and robust disturbance compensation. 
Future work will focus on integrating online trajectory planning and closed-loop control for extended aerial applications, including environmental monitoring and autonomous inspection.


\section*{Acknowledgments}
The authors would like to thank Prof. Zhongkui Li for his help with the motion capture experiment. 

%%%
\bibliographystyle{IEEEtran}
\bibliography{Reference}
%%%


\begin{IEEEbiography}[{\includegraphics[width=1in,height=1.25in,clip,keepaspectratio]{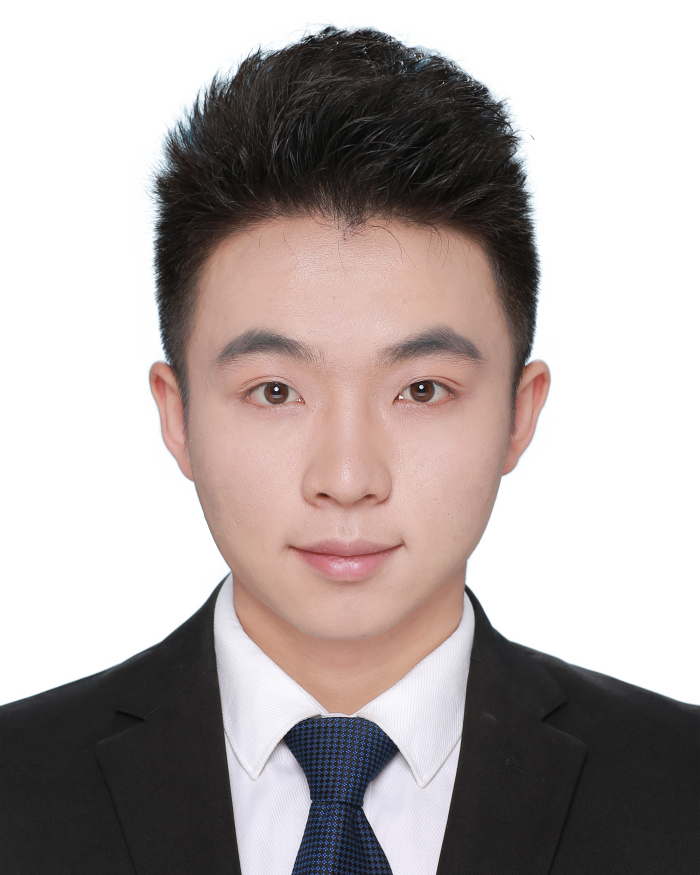}}]{Hao Cheng} (Student Member, IEEE) 
received the Bachelor's degree in New Energy Science and Engineering (Wind Power) from North China Electric Power University, China, in 2017, and the Master's degree in Control Engineering from Tsinghua University, China, in 2021. 
He is currently working toward the Ph.D. degree in Mechanics at Peking University, China, under the supervision of Prof. F. Zhang. 
His research interests include \mbox{lighter-than-air} aerial vehicles, bio-inspired robotics, and continuum robots. 
\end{IEEEbiography}


\begin{IEEEbiography}[{\includegraphics[width=1in,height=1.25in,clip,keepaspectratio]{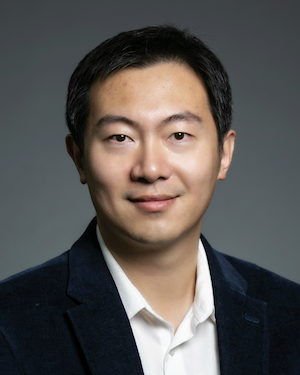}}]{Feitian Zhang} (Member, IEEE)
received the Bachelor's and Master's degrees in Automatic Control from Harbin Institute of Technology, Harbin, China, in 2007 and 2009, respectively, and the Ph.D. degree in Electrical and Computer Engineering from Michigan State University, East Lansing, MI, in 2014. 

He is currently an Associate Professor with the School of Advanced Manufacturing and Robotics, Peking University, Beijing, China. 
Prior to joining Peking University, he was an Assistant Professor in the Department of Electrical and Computer Engineering at George Mason University (GMU), Fairfax, VA, and the founding director of the Bioinspired Robotics and Intelligent Control Laboratory (BRICLab) from 2016 to 2021. 
His research interests include bioinspired robotics, control systems, artificial intelligence, underwater vehicles, and aerial vehicles. 

\end{IEEEbiography}

\vfill

\end{document}